\documentclass[runningheads]{llncs}
\usepackage[T1]{fontenc}
\usepackage[hidelinks]{hyperref}

\usepackage{graphicx}
\usepackage[dvipsnames]{xcolor}
\usepackage{times}
\usepackage{latexsym}
\usepackage{todonotes}
\usepackage{subcaption}
\usepackage[T1]{fontenc}
\usepackage{amsbsy}
\usepackage{amsmath}
\usepackage[utf8]{inputenc}

\usepackage{microtype}

\usepackage{inconsolata}
\usepackage{comment}

\begin{document}
\title{VerAs: Verify then Assess STEM Lab Reports}
%
%
\author{Berk Atil, Mahsa Sheikhi Karizaki \and
Rebecca J. Passonneau}
\authorrunning{Atil et al.}
%
\institute{The Pennsylvania State University, University Park 16801, USA\\
\email{\{bka5352,mfs6614,rjp49\}@psu.edu}
}
\maketitle              
\begin{abstract}
With an increasing focus in STEM education on critical thinking skills, science writing plays an ever more important role. 
A recently published dataset of two sets of college level lab reports from an inquiry-based physics curriculum relies on analytic assessment rubrics that utilize multiple dimensions, specifying subject matter knowledge and general components of good explanations. Each analytic dimension is assessed on a 6-point scale, to provide detailed feedback to students that can help them improve their science writing skills. Manual assessment can be slow, and difficult to calibrate for consistency across all students in large enrollment courses with many sections. While much work exists on automated assessment of open-ended questions in STEM subjects, there has been far less work on long-form writing such as lab reports. We present an end-to-end neural architecture that has separate verifier and assessment modules, inspired by approaches to Open Domain Question Answering (OpenQA). VerAs first verifies whether a report contains any content relevant to a given rubric dimension, and if so, assesses the relevant sentences. On the lab reports, VerAs outperforms multiple baselines based on OpenQA systems or Automated Essay Scoring (AES). VerAs also performs well on an analytic rubric for middle school physics essays.


\end{abstract}

\section{Introduction}
Science writing plays an important role in science education, whether to prepare students for science careers, or to nurture a more informed citizenry. Informative, reliable and timely feedback on written work supports learning \cite{evans2013making,o2016scholarly}, which in turn is often facilitated through rubrics. A recent meta-review of rubric usage throughout the educational cycle across different subject areas found a positive effect on student learning and performance~\cite{panadero-et-al2023}. Yet rubrics are time-consuming for educators to develop and use. Further, when teaching assistants (TAs) apply rubrics, the results can be unreliable \cite{becky_data}, reducing their benefit on learning. Automated support for assessment of writing has often addressed non-STEM automated essay scoring (AES; holistic scores)~\cite{bai-etal-2022,chen-li-2023-pmaes,do-etal-2023-prompt,wangEtAl-naacl22,xie-etal-2022-automated,yangEtAl-2020-enhancing}, or short answer assessment in STEM \cite{camusEtAl_transformer-asag_AIED20,condor2022representing,filigheraEtAl_biling-asag_acl22,liEtAl-2021-semantic,takano&ichikawa-2022-automatic,sungEtAl-2019-pre} or non-STEM~\cite{mizumoto-etal-2019-analytic,wang2021data}. 
There has been far less work on automated support to apply analytic rubrics for long-form STEM writing. Our work addresses automated application of analytic rubrics.

Panadero et al. \cite{panadero-et-al2023} define a rubric as setting expectations for student work through specification of evaluative criteria, and how to meet them. Their meta-review includes studies where rubrics lack a scoring strategy, as when the main goal is formative assessment, which occurs during a course while students are learning the material, to help them improve by the end of the course. A rubric is analytic if it specifies multiple criteria, or rubric dimensions.We designed an automated approach 
for rubric assessment of STEM writing, and evaluated it on college level physics lab reports that have a scoring strategy in the rubric, and on middle school essays where there is no scoring strategy.


Fig.~\ref{fig:rubric_dims} illustrates a key challenge with the lab report rubrics: 
they can use different scoring strategies. 
The top of the figure shows part of the first rubric dimension for lab reports on the behavior of a pendulum. This is a criterion-based rubric where each point increment requires more explanation and correctness. The bottom of the figure shows part of the seventh dimension of a rubric for a report on Newton's second law. Here an inclusion-based criterion is used, and the scoring strategy is to sum all the points.

A second challenge is that in discursive science writing, it can be difficult to localize what part of a report is relevant for a given rubric dimension.  As we discuss later in the paper, while human assessors can perform reliably on assigning a score for each dimension, they do not agree well on exactly which sentences address a given dimension.

\begin{figure}[t]
\small
    \centering
    \begin{tabular}{l|l}\hline
    \multicolumn{2}{l}{\textbf{Pendulum D1}: Is able to state the research question for  reader clarity} \\\hline
    \multicolumn{1}{c|}{Points} & \multicolumn{1}{c}{Select One}\\\hline
    1     &  Research question is included but incorrect. No mention of the three variables.\\
    . . . & . . .\\
    5     & Research question is included and correct:  \textit{What affects the period of a pendulum?} \\
          & Includes an explicit statement of the 3 variables: mass, angle of release, and string length.\\\hline
     \multicolumn{2}{l}{\textbf{Force \& Motion D7}: Is able to identify random errors and how they were or could be reduced. } \\\hline
     \multicolumn{1}{c|}{Points} & \multicolumn{1}{c}{Sum all that apply} \\\hline
      1   &  Discusses one random error.\\
       . . . & . . .\\
      1 & Includes one or more additional random or systematic errors. \\\hline 
    \end{tabular}
    \caption{A rubric dimension from each of two lab reports, with different scoring strategies.}
    \label{fig:rubric_dims}
\end{figure}

Given a rubric with $n$ dimensions and student lab reports, our assessment task is to generate a score for each dimension-report pair in the range  $[0:5]$. 
Inspired by Open-Domain Question Answering, we propose VerAs\footnote{The code for VerAs is available at \url{https://github.com/psunlpgroup/VerAs}}, which has a verifier module to determine whether a report contains sentences relevant to a dimension, and a grader to score the relevant sentences selected by the verifier. We test its effectiveness on a published dataset of lab reports~\cite{data}  
against multiple baselines. Through ablations, we demonstrate the need for both modules, and the benefit of using an ordinal loss training objective for the grader. We provide detailed error analysis of performance differences across rubric dimensions. To demonstrate the generality of the architecture, we also report results on middle school physics essays where the grader module is not necessary. We present related work, the datasets, VerAs architecture, experiments and results.


\section{Related Work}
\label{sec:related}

As noted in the introduction, AES and short answer assessment are active areas of research. In contrast, we find little work that attempts to automate rubric-based assessment for essays or lab reports. Ariely et al.~\cite{ariely2022machine} developed a method to detect biology concepts using convolutional neural networks in high school students' short explanation essays in Hebrew. Rahimi et al.~\cite{rahimiEtAl17} automated a rubric to assess students' use of evidence and organization of claims in source-based non-STEM writing. Ridley et al.~\cite{Ridley_He_Dai_Huang_Chen_2021} and Shibata \& Uto~\cite{shibata&uto-2022-analytic} present neural models that assess specific traits to support holistic scores on a widely used dataset of non-STEM argumentative, narrative, and source-dependent essays~\cite{mathias&bhattacharyya_asap_lrec18}. Apart from \cite{ariely2022machine}, our work differs in its focus on rubric criteria for specific explanatory content, e.g., about energy, periodicity in a pendulum, or force and motion.


Our approach 
is inspired by Open Domain Question Answering (OpenQA), where the goal is 
to query multiple documents, some of which may contain no relevant information. 
Most OpenQA systems have two modules, a retriever to find relevant sentences, and a reader to extract the answer \cite{hu2019read,izacard2021distilling,lee-etal-2019-latent,lee-etal-2018-ranking_new,sachanEtAl-2021-end,singh2021end}. %
Izacard \& Grave~\cite{izacard&grave-2021-leveraging} combine Dense Passage Retrieval (DPR) \cite{karpukhin-etal-2020-dense} with a sequence-to-sequence transformer reader module. In later work, they propose FiD-KD to perform knowledge distillation as a way to compensate for training data that lacks labeled pairs of queries and documents with answers~\cite{izacard2021distilling}.
Similarly, Read+Verify \cite{hu2019read+} has a distinct module to assess whether a question-passage pair can provide an answer.
VerAs processes sentences rather than passages, but also relies on a verifier module to first determine whether a lab report contains sentences relevant to a given rubric dimension. Similar to \cite{izacard2021distilling,hu2019read+}, we lack annotations on which sentences in a report, if any, are relevant to each rubric dimension.

\begin{table}[t]
\centering
\begin{tabular}{l|r|r|r|r|r|r} \hline
    \multicolumn{1}{c}{Split} & 
        \multicolumn{2}{|c}{Pendulum} & 
            \multicolumn{2}{|c}{Newton's 2nd Law} & 
                \multicolumn{2}{|c}{Both} \\\cline{2-7}
    & \multicolumn{1}{|c}{N} &            \multicolumn{1}{|c}{$Len_{sent}$} &
        \multicolumn{1}{|c}{N} &            \multicolumn{1}{|c}{$Len_{sent}$} &
            \multicolumn{1}{|c}{N} &         \multicolumn{1}{|c}{$Len_{sent}$} \\\hline
    Train & 868 & 25.47 (12.95)& 798 & 25.73 (12.88) & 1,666 & 25,59 (12.92) \\
    Val.  & 108 & 26.35 (13.57) & 101 & 26.01 (14.19) & 209 & 26,19 (13.88)\\
    Test  & 102 & 26.51 (15.80)& 106 & 25.48 (12.74)& 208 & 25,98 (14.33) \\\hline\hline
            \multicolumn{1}{c}{Split} & 
        \multicolumn{2}{|c}{Essay 1} & 
            \multicolumn{2}{|c}{Essay 2} & 
                \multicolumn{2}{|c}{Both} \\\cline{2-7}
    & \multicolumn{1}{|c}{N} &            \multicolumn{1}{|c}{$Len_{sent}$} &
        \multicolumn{1}{|c}{N} &            \multicolumn{1}{|c}{$Len_{sent}$} &
            \multicolumn{1}{|c}{N} &         \multicolumn{1}{|c}{$Len_{sent}$} \\\hline
        Train & 899 &16.35 (8.82) &720 & 21.60 (11.59) & 1619 & 18.69 (10.47) \\
        Val   & 95 & 14.27 (7.40) & 95 & 21.47 (12.48) & 190 & 17.87 (10.87) \\
        Test  & 99 & 19.15 (10.15) & 56 & 31.00 (21.53) & 155 & 23.43 (16.30) \\
        Total & 1,093  & & 871 & & 2,003 &\\\hline
\end{tabular}
\vspace{.1in}
\caption{The top five rows for the college lab reports, and the bottom five for the essays, give the count in each data split, and mean length (sd) in sentences.}
\label{tab:dataset_statistics}
\end{table}

\section{Datasets}

\begin{figure*}[t]
\centering
\subfloat[Dimension 6 in the second lab.]{%
    \fbox{\includegraphics[width={0.39\textwidth}]{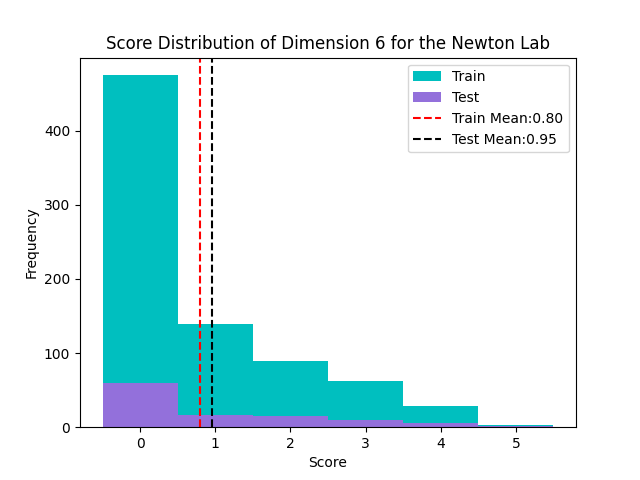}}}\hfil
\subfloat[All dimensions in the second lab.]{%
    \fbox{\includegraphics[width={0.37\textwidth}]
    {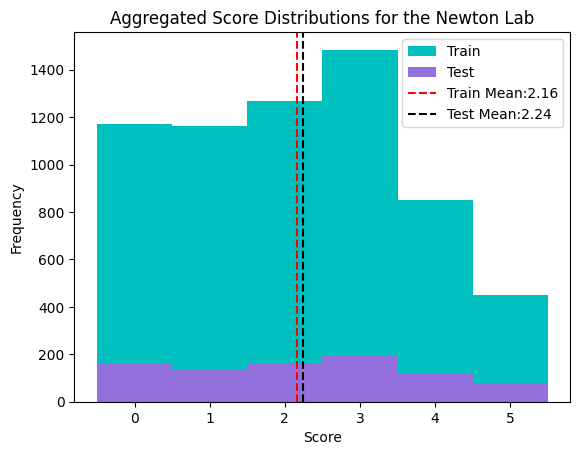}}}\hfil
\caption{For both lab reports, score distribution per dimension is highly skewed towards low or high scores, depending on the dimension difficulty, as in (a). The skew is less apparent when scores are aggregated across dimensions, as in (b).
}
\label{fig:Dimension_dist}
\end{figure*}

The college physics dataset consists of two sets of lab reports~\cite{data} from a curriculum designed to promote scientific reasoning skills. The first, about factors affecting the period of a pendulum, has a 7-dimension rubric. The second report, on Newton's Second Law, has an 8-dimension rubric. Each rubric dimension specifies precise criteria for each point increment on a six-point scoring scale, as illustrated in Fig.~\ref{fig:rubric_dims}; the supplemental provides the complete rubrics. Each report has a ground truth score for each dimension from one of four trained raters. On random subsets of multiply labeled reports, raters had an average Pearson correlation of 0.72 on the 7 dimensions of the first report, and 0.69 on the 8 dimensions of the second report. The top half of Table~\ref{tab:dataset_statistics} shows the size of the dataset splits (training, validation, test) and mean sentence lengths. As shown in Fig. \ref{fig:Dimension_dist}, scores per dimension are highly skewed. 

The middle school data consists of responses to two essay prompts from a unit on the physics of roller coasters~\cite{puntambekarEtAl_science-explanations_icls23}, as shown in the bottom half of Table~\ref{tab:dataset_statistics}. The first essay rubric identifies six main ideas about energy and the law of conservation of energy. The second essay rubric adds two additional ideas about the relations of mass to speed, and height to speed. Only 159 of the essays have reliable manual labels indicating the presence 
of main ideas (Cohen's kappa = 0.77) (essay 1 test is entirely manual labels). The remaining labels are from an automated tool called PyrEval~\cite{gao-etal-2019-automated,singh_automated_2022} whose accuracies on the two essays are 0.76 and 0.80, respectively, as reported below. 
For essay 2, there are reliable manual labels on 56 essays, corresponding to the essay 2 test set.

\section{VerAs Task and Architecture}
\label{sec:veras-architecture}

\begin{figure*}[t]
    \centering
    \includegraphics[width=\textwidth]{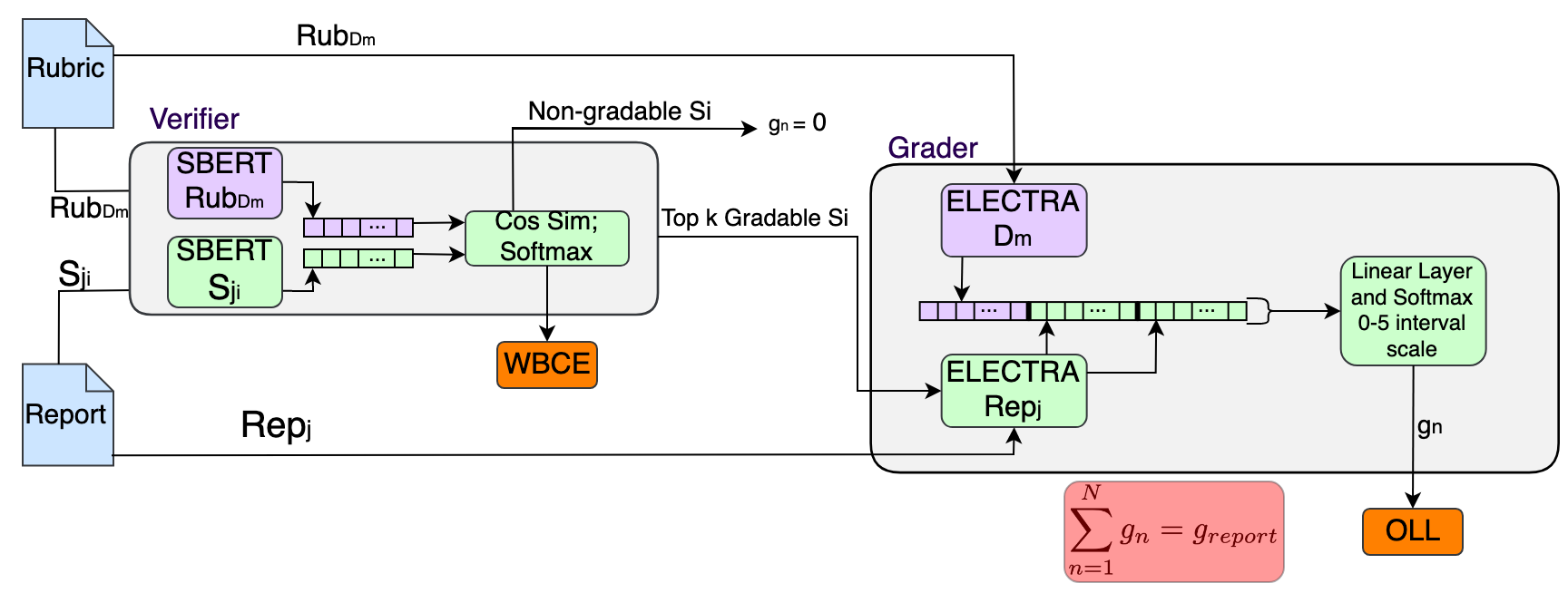}
    \caption{VerAs: Using a dual encoder, the verifier assesses each report sentence ($S_i$) and rubric dimension ($D_m$) to forward the top $k$ sentences to the grader, trained with weighted binary cross-entropy loss on whether the report receives a non-zero score. The grader also uses a dual encoder; it concatenates the top $k$ sentences, $D_m$, and the full report $Rep_j$, trained with ordinal log loss as the training objective to assign a score.
    }
    \label{fig:veras-arch}
\end{figure*}

VerAs treats each dimension of a rubric as a query, where the response to each query is a score in $[0:5]$. We make the simplifying assumption that at most a few sentences of a report are relevant for the assessment of a given dimension. To address the challenge that we lack labels on which sentences are relevant, we developed a pipeline with one module to select relevant sentences, and a subsequent module to apply the score. To address the challenge of the diversity and complexity of the dimensions (cf. Fig. \ref{fig:rubric_dims}), each module has a dual encoder to learn better similarities of sentences to dimensions.
As illustrated in Fig.~\ref{fig:veras-arch}, each sentence in a report is paired with each dimension and passed to the verifier, which in turn passes relevant sentences, the dimension, and the full report to the grader. The next two sections describe the verifier and grader in detail.

To develop a better understanding of the difficulty of the sentence selection process, the first author and a colleague independently selected relevant sentences for each dimension of 20 lab reports (10 from each of the two assignments), with no constraint on how many sentences to select. Both raters had access to the ground truth score on each dimension. We assess their agreement using Krippendorff's alpha combined with a distance metric developed for comparison of raters on set selection tasks~\cite{passonneau-masi-lrec06}.  Depending on the dimension and rater, the average number of selected sentences ranged from 1.3 to 8.4 ($\mu=3.4, \sigma=1.5$). Rater agreement on lab one was 0.54 and 0.43 on lab two. Thus the task is difficult for humans to achieve with consistency, and humans vary greatly in the number of relevant sentences they select. We attribute this in part to many sentences having multiple clauses where only part of a sentence might be relevant.

\subsection{Verifier}

The verifier makes two decisions: deciding if a report should receive a non-zero score on a given dimension, and if so, determining which sentences are the most relevant. Because we only have labels on the first decision, and because the data is imbalanced (see Fig.~\ref{fig:rubric_dims}), we use weighted binary cross-entropy as the loss function.  To find relevant sentences, we learn representations for the sentences and rubric dimension to achieve meaningful similarity. 
Let the sentences in a lab report be denoted by $S=\{s_1,s_2,...,s_n\}$. Given a rubric dimension $q$, we calculate the cosine similarity between the embeddings of the rubric dimension and each sentence as follows:
\begin{align}
cos\_sim(q,s;\theta_q, \theta_s) = 
\frac{f(q;\theta_q)^T f(s;\theta_s)}{max(||f(q;\theta_q)||_2 ||f(s;\theta_s)||_2, \epsilon)}
\label{equation:verifier}
\end{align}

\noindent
where $\theta_q$ and $\theta_s$ are the parameters of our encoder functions for the rubric dimensions and report sentences respectively, and $\epsilon$ is a small value to avoid division by 0. A dual encoder \cite{dual_encoder} learns different embedding spaces for the rubric dimensions versus report sentences, using SBERT \cite{reimers-gurevych-2019-sentence}, which was designed to learn representations for semantic similarity comparisons more efficiently.  
After the calculation of pairwise cosine similarities, the top $k$ similarities are averaged and converted to a probability using equation \ref{equation:softmax}, as in \cite{bridle1989training}:
\begin{align}
f_{softmax} = \frac{1}{1+e^{-10(D-0.5)}}
\label{equation:softmax}
\end{align}
\noindent
where $D$ is the mean of the top $k$ cosine similarity values. 


\subsection{Grader}

Similar to the verifier, the grader relies on a dual encoder to learn more effective similarities of the encoded report ($r$) and top $k$ relevant sentences ($rel$) with the rubric dimension. Inclusion of $r$ provides a global context for $rel$, and potentially compensates for the possibility $rel$ fails to include all the relevant sentences. This is likely given that $k$ is fixed once VerAs is trained, whereas we found high variability in the number of sentences that human raters selected as relevant, across dimensions and reports. 

The grader calculates the probability distribution $P$ over scores (six classes) as:
\begin{align}
P(q,r,rel;\beta_q, \beta_r, \phi)  =  
 f_{softmax}(f([g(q;\beta_q), g(r;\beta_r), g(rel;\beta_r)];\phi))
\label{equation:grader}
\end{align}
\noindent
where $f$ is a linear layer, $g$ is the encoder, and $\beta_q$, $\beta_r$, and  $\phi$ are the learned parameters of the encoders for the rubric dimension, report and prediction layers, respectively. 
We experiment with BERT \cite{devlin_bert_naacl19} 
ELECTRA \cite{Clark2020ELECTRA} and 
LongT5 \cite{guo-etal-2022-longt5} for the encoder function. 

Cross entropy loss (CE) is not appropriate for our task, because the score classes are on an ordinal scale where the distance between pairs of values varies. 
Therefore, we use ordinal log loss (OLL) \cite{castagnos-etal-2022-simple} as the grader's loss function: 

\begin{equation}
 L_{OLL\mbox{-}\alpha}(P, y) = -\sum_{i=1}^N \log(1-p_i) \delta(y,i)^{\alpha}
\label{equation:oll}
\end{equation}
given $N$ classes, $P$ as the model's estimated probability distribution, the true label $y$, a distance function $\delta$,  and a hyperparameter $\alpha$. For $\delta$, we use absolute distance.

\section{Experiments}

Experiments on the lab reports compare VerAs with multiple baseline models, plus the majority class baseline. We also perform five VerAs ablations to assess its components. We test only the VerAs verifier module on the essays, as explained further below.

All experiments use the Adam optimizer \cite{kingma2014adam}, and the same learning rates (0.001, 0.0001, 0.00001, 0.005, 0.0005, 0.00005) and batch sizes (4, 8, 16). We select the optimal hyperparameters given the validation loss, except for  R$^2$BERT. Its loss automatically decreases each epoch because of its dynamic weight strategy, so we rely on the Spearman correlation instead. We tune $\alpha$ for  OLL with $1, 1.5, 2, 2.5,$ and $3$. Lastly, we try $1, 2, 3, 4, 20,$ and $25$ as the top $k$ parameter for both VerAs and FiD-KD. For FiD-KD, 25 is best, which is close to the average report length. For VerAs, 3 is best, which is close to the average of our two human raters (see section~\ref{sec:veras-architecture}).

\subsection{Baselines}

Two of the baselines are distinct models, and one is a variant of VerAs. The R$^2$BERT \cite{yangEtAl-2020-enhancing} AES system predicts a total score given a report, without utilizing the rubric dimensions. It uses a BERT encoder followed by a linear layer to predict the score, scaling the scores to $[0-1]$. The loss is a dynamically weighted sum of a regression (MSE) and ranking loss (CE). We tune the learning rate, batch size, and truncation size of the report. 


The second baseline is the OpenQA system that most directly inspires VerAs, FiD-KD \cite{izacard2021distilling} (see above). 
In the reader module, T5 encodes the question and passage. The concatenation of their vectors goes to the T5 decoder. The retriever uses a BERT-based bi-encoder to assess the similarity of a question-passage pair, similar to DPR~\cite{karpukhin-etal-2020-dense}. Knowledge distillation is performed from the reader to the retriever, using reader attention scores as pseudo-labels to train the retriever. Here, we treat each rubric dimension as a question and each sentence as the passage. The class names ($[0:5]$) are spelled out, and a single time step decoding is carried out on this restricted vocabulary, as in \cite{schick2021self}.  


The third baseline reimplements VerAs as a multi-task model: each rubric dimension becomes a separate problem, with a separate classification layer in the grader for each rubric dimension. The verifier module remains the same. VerAs$_{SEP}$ thus tests whether different classifiers are needed to handle the semantic diversity across rubric dimensions.





\subsection{Ablations}

The first ablation replaces the verifier with random selection of three sentences from each lab report (\textit{Random Verifier}). The second and third ablations omit the verifier module altogether, with the grader receiving only the rubric dimension and report, using either a truncated report to meet the input length constraint (\textit{W/o Verifier Trunc.}), or an average of embeddings of a moving window over the full report (\textit{W/o Verifier Mov. Avg.}). In the fourth ablation, the input to the grader omits the report (\textit{W/o Report}). 
The final ablation, VerAs$_{CE}$, uses cross entropy loss instead of OLL.

\section{Results}
We evaluate the performance of VerAs on the lab reports in two ways: 
on the total report score, which is the sum of the scores on each dimension, and also at the dimension level. In this section, we first present the evaluation metrics used here, then the two types of results, followed by error analysis. The final subsection presents results of the verifier module on the middle school physics essays.

\subsection{Evaluation Metrics}

The total score on a report is the sum of the scores on each dimension. To evaluate the total score, we report Mean Squared Error (MSE), Krippendorff’s alpha coefficient ($\alpha_{Interval}$), and weighted accuracy. MSE is the squared difference between the prediction and ground truth. Agreement coefficients like Krippendorff's alpha \cite{krippendorff11}, which factor out agreements that could arise by chance, are most familiar with categorical decisions but use of an interval scale supports comparison of two numeric outcomes. Similarly, weighted accuracy takes the absolute distance between the ground truth and prediction into account; it is calculated as follows:

\begin{equation}
W_{acc} = \frac{\sum_{i=1}^n \frac{1 - |g_n^i-y_n^i|}{\textit{max distance}}}{n}
\label{equation:weighted_acc}
\end{equation}
where $g_n^i$ and $y_n^i$ are the prediction and ground truth for the $n$th rubric dimension of report $i$, and max distance is the maximum absolute difference between the prediction and ground truth: 5 for the dimension level, 35 for the first lab, and 40 for the second.

The predicted total score could be correct without being correct on any one dimension, so we also evaluate how well the scores on each dimension agree. We report the Spearman correlation, which measures the distance between two rankings, by averaging the Spearman correlations of the per dimension predictions with the ground truth over all reports. We also report the average $\alpha_{Interval}$.

For the verifier, we evaluate its decision as to whether a report gets a non-zero grade, using accuracy, micro-averaged precision, recall and F1-score.

\subsection{Results by Total Score and by Dimension}

Table \ref{tab:holistic_results} shows that VerAs outperforms all of the baselines: by at least 17.9\% on MSE, 8.0\% on $\alpha_{Interval}$, and 0.8\% on weighted accuracy. On total score, VerAs$_{SEP}$ performs less well than VerAs, possibly because each classifier has only 1,666 examples instead of 12,460. Surprisingly, R$^2$BERT outperforms FiD-KD in two metrics although it uses a simpler architecture. VerAs also outperforms the ablations on MSE and $\alpha_{interval}$, especially when CE instead of OLL is used. The weighted accuracy results are uniformly high due to the extreme data skew, but show no sensitivity across models. Table \ref{tab:holistic_correlation}, which gives the average per dimension correlations and agreement, shows VerAs$_{SEP}$ to have the highest performance, with VerAs outperforming FiD-KD.


\begin{table*}[t!]
\centering
\setlength{\tabcolsep}{8pt}
   \begin{tabular}{l|c|c|c}
        \hline
        \multicolumn{1}{c|}{\textbf{Model} }  & \multicolumn{1}{c|}{\textbf{MSE}} & \multicolumn{1}{c|}{\textbf{\pmb{$\alpha_{Interval}$}}} & \multicolumn{1}{c}{\textbf{Weighted Acc.}}   \\ 
        \hline
        \multicolumn{4}{c}{Comparison with baselines}  \\
        \hline
        VerAs &  \textbf{19.11 (19.09, 19.13)}  & \textbf{0.77 (0.77, 0.77)} & \textbf{0.91 (0.91, 0.91)}  \\
        VerAs$_{SEP}$ & 23.27 (23.25, 23.29)  & 0.70 (0.70, 0.70)  & 0.90 (0.90, 0.90) \\
         R$^2$BERT  & 27.05 (27.03, 27.07) & 0.68 (0.68, 0.68)  &  0.89 (0.89, 0.89)\\ 
         FiD-KD &  27.46 (27.43, 27.48) & 0.67 (0.67, 0.67)  & 0.89 (0.89, 0.89) \\ 
          \hline
         \multicolumn{4}{c}{Ablations}  \\
         \hline
          Random Verifier  & 19.24 (19.23, 19.26)  & 0.69 (0.69, 0.69) & \textbf{0.91 (0.91, 0.91)}  \\
         W/o Verifier Trunc.  & 19.16 (19.14, 19.17) & 0.67 (0.67, 0.67)  & \textbf{0.91 (0.91, 0.91)} \\
         W/o Verifier Mov. Avg. &  20.65 (20.64, 20.67) & 0.68 (0.68, 0.68) & 0.90 (0.90, 0.90)\\
         W/o Report  &  20.85(20.83, 20.87) & 0.70 (0.70,0.70) & \textbf{0.91 (0.91, 0.91)} \\
        VerAs$_{CE}$  & 24.29 (24.27, 24.32) &  0.71 (0.71, 0.71)  & 0.89 (0.89, 0.89) \\
        \hline
        \end{tabular}
        
\caption{Total report score evaluations with 95\% bootstrapped confidence intervals.}
\label{tab:holistic_results}
\end{table*}

\begin{table*}[t!]
\centering
   \begin{tabular}{l|c|c|c|c}
        \hline
        \multicolumn{1}{c|}{\textbf{Model} }  & \multicolumn{1}{c|}{\textbf{Spearman Pend.}} & \multicolumn{1}{c|}{\textbf{\pmb{$\alpha_{Interval}$} Pend.}} & \multicolumn{1}{c|}{\textbf{Spearman Newt.}} & \multicolumn{1}{c}{\textbf{\pmb{$\alpha_{Interval}$} Newt.}}   \\ 
        \hline
        \multicolumn{5}{c}{Comparison with baselines}  \\
        \hline
        VerAs & 0.52 (0.36) & 0.46 (0.35)  & 0.60 (0.30) & \textbf{0.54 (0.30)} \\
        VerAs$_{SEP}$ & \textbf{0.59 (0.33)} & \textbf{0.53 (0.36)} &  \textbf{0.62 (0.27)} & \textbf{0.54 (0.28)} \\
         FiD-KD & 0.53 (0.36) & 0.46 (0.36)  &   0.49 (0.37) & 0.41 (0.32) \\ 
          \hline
         \multicolumn{5}{c}{Ablations}  \\
         \hline
         Random Verifier  & 0.45 (0.38) & 0.27 (0.31)  &  0.48 (0.32) & 0.35 (0.26)   \\
         W/o Verifier Trunc. & 0.45 (0.39) & 0.29 (0.32) &  0.49 (0.32) & 0.38 (0.26)  \\
         W/o Verifier Mov. Avg. & 0.44 (0.37) & 0.28 (0.30)   &  0.48 (0.34) & 0.35 (0.27) \\
         W/o Report  &  0.49 (0.34) & 0.40(0.33) & 0.58 (0.31) & 0.52 (0.32)  \\
        VerAs$_{CE}$  & 0.42 (0.41) &0.33 (0.38) &  0.44 (0.31)  & 0.37 (0.30) \\\hline
        \end{tabular}        
\caption{Average correlations across dimensions for each lab, along with the mean (std).}
\label{tab:holistic_correlation}
\end{table*}

\subsection{Error Analysis of the Verifier's Binary Decision}

With respect to overall performance, Table \ref{tab:verifier_results} shows that the verifier does a better job on lab 1, which is also easier for the students: 
Fig. \ref{fig:Dimension_dist}b) shows the mean score on lab 2 to be 2.16 in the training data, compared to 2.95 on lab 1, but with relatively few zero scores on any dimension (see supplemental). We speculate that the verifier does better on lab 1 because the data is more balanced. On each dimension, verifier accuracy is often close to the majority class baseline. However, for dimension 6 on lab 1, and dimensions 2-5 on lab 2, it is lower than the majority class result; for dimensions 1, 6 and 8 on lab 2, the verifier accuracy is greater than the majority class baseline. In general, it 
provides good sentences, which is the more important responsibility of the verifier and through ablation studies, we show its effectiveness.
There appears to be a relationship between the difficulty of the rubric dimension and the performance of the verifier for the second lab. We calculate the pearson and spearman correlations between the accuracy of the verifier and the average training ground truth scores for each rubric dimension and we get 0.96 and 0.78 respectively. However, there is no such correlation for the first lab.
\begin{table*}[hbt!]
   \centering
    \begin{tabular}{l|c|c|c|c|c |c| l|c|c|c|c|c} \hline 
    \multicolumn{6}{c}{Pendulum} & 
        \multicolumn{1}{|c}{} & 
            \multicolumn{6}{|c}{F \& M} \\ \cline{1-6} \cline{8-13}
        \multicolumn{1}{l|}{\textbf{Dim.} } & \multicolumn{1}{c|}
        {\textbf{Maj. Base.}}
        & \multicolumn{1}{c|}{\textbf{Acc.}} & \multicolumn{1}{c|}{\textbf{Prec.}} & \multicolumn{1}{c|}{\textbf{Rec.}} & \multicolumn{1}{c|}{\textbf{F1} } & \multicolumn{1}{c|}{\text{    }} &
        \multicolumn{1}{l|}{\textbf{Dim.} } & \multicolumn{1}{c|}
        {\textbf{Maj. Base.}}
        & \multicolumn{1}{c|}{\textbf{Acc.}} & \multicolumn{1}{c|}{\textbf{Prec.}} & \multicolumn{1}{c|}{\textbf{Rec.}} & \multicolumn{1}{c}{\textbf{F1} } \\ \cline{1-6} \cline{8-13}
        1 & 0.90 & 0.90 & {0.96} & 0.93& 0.95 &  & 1 &0.92 &{0.95} & 0.98 & {0.97} & {0.97} \\
         2 & 0.95 & 0.96 & 0.97& 0.99& 0.98 & & 2 & 0.88 & 0.82 & {1.00} & 0.80 & 0.89\\ 
         3 & 0.98 & 0.99 & 0.99 & {1.00}& {1.00}  & &  3 & 0.90 & 0.84 & 0.98 & 0.84 & 0.90\\ 
         4 & 0.96 & 0.97 & 0.98& 0.99 & 0.98  & & 4 & 0.85 & 0.75 & 0.97 & 0.72 & 0.83\\ 
         5 & 0.93 & 0.93 & {1.00} & 0.93 & 0.96  & & 5 & 0.84 & 0.82 & 0.96 & 0.82 & 0.88\\ 
         6 & 0.98 &0.94 & {1.00}&  0.94& 0.97  & & 6 & 0.56 & {0.61} & {1.00} & {0.13} & {0.23}\\ 
         7 & 0.76 & 0.75 & {0.96} & {0.71} & {0.81}  & & 7 & 0.92 & 0.92 & 0.97 & 0.94 & 0.95\\ 
           & &  & &  &  & & 8 & 0.76 & 0.80 & 0.98 & 0.75 & 0.85 \\\hline
         Overall & 0.92 & 0.92 & 0.91 & 0.92 & 0.91  & & Overall & 0.81 & 0.81 & 0.83 & 0.81 & 0.80 \\ \hline
\end{tabular}
\vspace*{.1in}
 \caption{Verifier binary decision scores for the first (left) and second (right) lab.} 
\label{tab:verifier_results}

\end{table*}
\begin{table*}[hbt!]
\centering
\setlength{\tabcolsep}{8pt}
\begin{tabular}{l|c|c|c } \hline 
\multicolumn{4}{c}{Essay 1} \\\hline
        \multicolumn{1}{l|}{\textbf{Idea} } & \multicolumn{1}{c|}{\textbf{VerAs Verifier}} & \multicolumn{1}{c|}{\textbf{FiD-KD}} &  \multicolumn{1}{c}{\textbf{PyrEval}} \\\hline
         1 & 0.68 (0.68, 0.69) & 0.67 (0.67, 0.67) & 0.65 (0.65, 0.65) \\ 
         2 & 0.62 (0.61, 0.62) & 0.70 (0.70, 0.70) & 0.66 (0.65, 0.66) \\
         3 & 0.67 (0.67, 0.67)  & 0.68 (0.68, 0.68) & 0.69 (0.69, 0.69) \\
         4 & 0.92 (0.91, 0.92) & 0.95 (0.95, 0.95) & 0.92 (0.91, 0.92) \\
         5 & 0.85 (0.85, 0.86) & 0.80 (0.80, 0.80)  & 0.85 (0.85, 0.86) \\
         6 & 0.81 (0.81, 0.82)  & 0.78 (0.78, 0.79)  & 0.81 (0.81, 0.82) \\\hline
         Overall & 0.76 (0.76, 0.76) & 0.76 (0.76, 0.77) & 0.76 (0.76, 0.76)\\\hline  
        \multicolumn{4}{c}{Essay 2} \\\hline
        \multicolumn{1}{l|}{\textbf{Idea} } & \multicolumn{1}{c|}{\textbf{VerAs Verifier}} & \multicolumn{1}{c|}{\textbf{FiD-KD}} &  \multicolumn{1}{c}{\textbf{PyrEval}} \\\hline
        1 & 0.87 (0.87, 0.87) & 0.84 (0.83, 0.84) & 0.82 (0.82, 0.82)\\ 
        2 & 0.93 (0.93, 0.93) & 0.93 (0.93, 0.93) & 0.93 (0.93, 0.93)  \\
        3 & 0.75 (0.74, 0.75) & 0.73 (0.72, 0.73) &  0.82 (0.82, 0.82)\\
        4 & 0.93 (0.93, 0.93) &  0.93 (0.93, 0.93) & 0.93 (0.92, 0.93)\\
        5 & 0.77 (0.76, 0.77) & 0.82 (0.82, 0.82) & 0.84 (0.83, 0.84)\\
        6 & 0.80 (0.80, 0.81)&  0.84 (0.84, 0.84) & 0.77 (0.77, 0.77)\\
        7 & 0.62 (0.62, 0.63) & 0.57 (0.57, 0.57) & 0.55 (0.55, 0.55) \\
        8 & 0.73 (0.73, 0.73) &  0.59 (0.59, 0.59) & 0.78 (0.78, 0.79) \\\hline
        Overall & 0.80 (0.80, 0.80)  & 0.78 (0.78, 0.78) & 0.80 (0.80, 0.81)\\ \hline  
        \end{tabular}        
\caption{Accuracies with confidence intervals on middle school essays. }
\label{tab:middle_school_results}
\end{table*}

\subsection{Results on Middle School Essays}

Like the lab report dataset, we have data for two middle school essay assignments, along with analytic rubrics for formative feedback, and where each rubric has a different number of dimensions (six for essay 1; eight for essay 2). Instead of dimensions that differ with respect to different aspects of an experiment, such as the research question, theoretical equation, or sources of error, each essay rubric dimension is an explanatory statement of one of the main ideas in the curriculum. These can be more general, such as how potential and kinetic energy in a roller coaster are related to one another, or more specific, such as an explanation of the law of conservation of energy. Instead of assessing each dimension on a scale, the essay feedback indicates only whether the student included a clear statement of one of the main ideas. As a result, the VerAs grader module plays no role. We include results of FiD-KD, and PyrEval.

PyrEval is a toolkit for assessing the content of short passages. From a small set of $N$ reference passages it can automatically create a content model, called a pyramid, which is then used to detect similar content in unseen passages, all written to the same prompt. Content units (CUs) in the pyramid are sets of paraphrases extracted from the reference passages, where each CU has an importance weight equivalent to the number of reference passages that expressed that content.  PyrEval can create content models from as few as 4 or 5 reference passages, and requires no training data. 

Table~\ref{tab:middle_school_results} shows that all three models have the same overall accuracy on essay 1, while FiD-KD has slightly lower accuracy on essay 2. The per-dimension accuracies differ only slightly across models, and follow the same trend lines.

\section{Conclusion}

Our results show that formative assessment of longer forms of student writing, even those as complex as college-level lab reports with very detailed rubrics, can be handled by a neural network. VerAs performs very well on two sets of college level lab reports at applying a fine-grained analytic rubric, outperforming strong baselines. Ablations show that omitting the verifier module lowers MSE and $\alpha_{Interval}$ on the total report score. This indicates the verifier plays an important role despite the lack of labeled data for the verifier sentence selection task. Evaluation of how well each dimension is scored, however, shows that VerAs$_{SEP}$ outperforms VerAs. On a less complex essay dataset, VerAs, FiD-KD and a content assessment toolkit that requires no training perform equally well. Future work might focus on incorporating the score definitions in the rubrics and a better strategy to deal with the lack of labeled data for the sentence selection task. Additionally, large language models such as GPT-4 \cite{openaigpt4} 
can be prompted to have potentially noisy labels for the relevant sentences. \\
\textbf{Limitations:} VerAs needs to be retrained for new datasets, 
which reduces its generality. Future work might focus on this by using large language models. We test VerAs performance on two college physics lab reports and one middle school physics essay. Future work might test on other STEM domains such as biology.

\section{Acknowledgements}
We thank Sarkar Das, Vipul Gupta, Zhaohui Li, and Ruihao Pan for helpful discussions. The second author's work was supported by NSF DRK award 2010351.

\bibliographystyle{splncs04}
\bibliography{custom,anthology}

\begin{thebibliography}{10}
\providecommand{\url}[1]{\texttt{#1}}
\providecommand{\urlprefix}{URL }
\providecommand{\doi}[1]{https://doi.org/#1}

\bibitem{openaigpt4}
Achiam, J.e.: {GPT-4} technical report (2024), arXiv 2303.08774

\bibitem{ariely2022machine}
Ariely, M., Nazaretsky, T., Alexandron, G.: Machine learning and {H}ebrew {NLP} for automated assessment of open-ended questions in biology. International journal of artificial intelligence in education pp. 1--34 (2022), \url{https://link.springer.com/article/10.1007/s40593-021-00283-x}

\bibitem{bai-etal-2022}
Bai, H., Huang, Z., Hao, A., Hui, S.C.: Gated character-aware convolutional neural network for effective automated essay scoring. In: IEEE/WIC/ACM Inter. Conf. on Web Intelligence and Intelligent Agent Technology. p. 351–359. {ACM} (2022). \doi{10.1145/3486622.3493945}

\bibitem{bridle1989training}
Bridle, J.: Training stochastic model recognition algorithms as networks can lead to maximum mutual information estimation of parameters. NIPS  \textbf{2} (1989)

\bibitem{dual_encoder}
Bromley, J., Guyon, I., LeCun, Y., S\"{a}ckinger, E., Shah, R.: Signature verification using a "{S}iamese" time delay neural network. In: Proceedings of the 6th Interntl. Conf. on Neural Information Processing Systems. p. 737–744. Morgan Kaufmann, San Francisco, CA (1993)

\bibitem{camusEtAl_transformer-asag_AIED20}
Camus, L., Filighera, A.: Investigating transformers for automatic short answer grading. In: International Conf. on Artificial Intelligence in Education ({AIED}). p. 43–48 (2020). \doi{10.1007/978-3-030-52240-7_8}

\bibitem{castagnos-etal-2022-simple}
Castagnos, F., Mihelich, M., Dognin, C.: A simple log-based loss function for ordinal text classification. In: Proceedings of the 29th International Conference on Computational Linguistics. pp. 4604--4609. International Committee on Computational Linguistics, Gyeongju, Republic of Korea (Oct 2022), \url{https://aclanthology.org/2022.coling-1.407}

\bibitem{chen-li-2023-pmaes}
Chen, Y., Li, X.: {PMAES}: Prompt-mapping contrastive learning for cross-prompt automated essay scoring. In: Proceedings of the 61st Annual Meeting of the Association for Computational Linguistics (Volume 1: Long Papers). pp. 1489--1503. Association for Computational Linguistics, Toronto, Canada (Jul 2023). \doi{10.18653/v1/2023.acl-long.83}

\bibitem{Clark2020ELECTRA}
Clark, K., Luong, M.T., Le, Q.V., Manning, C.D.: {ELECTRA}: Pre-training text encoders as discriminators rather than generators. In: International Conference on Learning Representations (2020), \url{https://openreview.net/forum?id=r1xMH1BtvB}

\bibitem{condor2022representing}
Condor, A., Pardos, Z., Linn, M.: Representing scoring rubrics as graphs for automatic short answer grading. In: Artificial Intelligence in Education: 23rd International Conference, AIED 2022, Durham, UK, July 27--31, 2022, Proceedings, Part I. pp. 354--365. Springer (2022)

\bibitem{devlin_bert_naacl19}
Devlin, J., Chang, M.W., Lee, K., Toutanova, K.: {BERT}: Pre-training of deep bidirectional transformers for language understanding. In: Burstein, J., Doran, C., Solorio, T. (eds.) Proceedings of the 2019 {NAACL} and {HLT}. pp. 4171--4186. Association for Computational Linguistics, Minneapolis, Minnesota (Jun 2019). \doi{10.18653/v1/N19-1423}

\bibitem{do-etal-2023-prompt}
Do, H., Kim, Y., Lee, G.G.: Prompt- and trait relation-aware cross-prompt essay trait scoring. In: Findings of the Association for Computational Linguistics: ACL 2023. pp. 1538--1551. Association for Computational Linguistics, Toronto, Canada (Jul 2023). \doi{10.18653/v1/2023.findings-acl.98}

\bibitem{evans2013making}
Evans, C.: Making sense of assessment feedback in higher education. Review of educational research  \textbf{83}(1),  70--120 (2013)

\bibitem{filigheraEtAl_biling-asag_acl22}
Filighera, A., Parihar, S., Steuer, T., Meuser, T., Ochs, S.: Your answer is incorrect... would you like to know why? introducing a bilingual short answer feedback dataset. In: Muresan, S., Nakov, P., Villavicencio, A. (eds.) Proceedings of the 60th {ACL}. pp. 8577--8591. Association for Computational Linguistics, Dublin, Ireland (May 2022). \doi{10.18653/v1/2022.acl-long.587}

\bibitem{gao-etal-2019-automated}
Gao, Y., Sun, C., Passonneau, R.J.: Automated pyramid summarization evaluation. In: Proceedings of the 23rd Conference on Computational Natural Language Learning (CoNLL). pp. 404--418. Association for Computational Linguistics, Hong Kong, China (Nov 2019). \doi{10.18653/v1/K19-1038}, \url{https://aclanthology.org/K19-1038}

\bibitem{guo-etal-2022-longt5}
Guo, M., Ainslie, J., Uthus, D., Ontanon, S., Ni, J., Sung, Y.H., Yang, Y.: {L}ong{T}5: {E}fficient text-to-text transformer for long sequences. In: Findings of the Association for Computational Linguistics: NAACL 2022. pp. 724--736. Association for Computational Linguistics, Seattle, United States (Jul 2022). \doi{10.18653/v1/2022.findings-naacl.55}, \url{https://aclanthology.org/2022.findings-naacl.55}

\bibitem{hu2019read}
Hu, M., Wei, F., Peng, Y., Huang, Z., Yang, N., Li, D.: Read+ verify: Machine reading comprehension with unanswerable questions. In: Proceedings of the AAAI Conference on Artificial Intelligence. vol.~33, pp. 6529--6537 (2019)

\bibitem{hu2019read+}
Hu, M., Wei, F., Peng, Y., Huang, Z., Yang, N., Li, D.: Read+ verify: Machine reading comprehension with unanswerable questions. In: Proceedings of the AAAI Conference on Artificial Intelligence. vol.~33, pp. 6529--6537 (2019)

\bibitem{izacard2021distilling}
Izacard, G., Grave, E.: Distilling knowledge from reader to retriever for question answering. In: ICLR (2021), \url{https://openreview.net/forum?id=NTEz-6wysdb}

\bibitem{izacard&grave-2021-leveraging}
Izacard, G., Grave, E.: Leveraging passage retrieval with generative models for open domain question answering. In: Merlo, P., Tiedemann, J., Tsarfaty, R. (eds.) Proceedings of the 16th EACL. pp. 874--880. Association for Computational Linguistics, Online (Apr 2021). \doi{10.18653/v1/2021.eacl-main.74}

\bibitem{karpukhin-etal-2020-dense}
Karpukhin, V., Oguz, B., Min, S., Lewis, P., Wu, L., Edunov, S., Chen, D., Yih, W.t.: Dense passage retrieval for open-domain question answering. In: Proceedings of the 2020 Conference on Empirical Methods in Natural Language Processing (EMNLP). pp. 6769--6781. Association for Computational Linguistics, Online (Nov 2020). \doi{10.18653/v1/2020.emnlp-main.550}, \url{https://aclanthology.org/2020.emnlp-main.550}

\bibitem{kingma2014adam}
Kingma, D.P., Ba, J.: Adam: A method for stochastic optimization. arXiv preprint arXiv:1412.6980  (2014)

\bibitem{krippendorff11}
Krippendorff, K.: Computing {K}rippendorff's alpha-reliability (2011), university of Pennsylvania Scholarly Commons, \url{https://repository.upenn.edu/asc\_papers/43}

\bibitem{lee-etal-2018-ranking_new}
Lee, J., Yun, S., Kim, H., Ko, M., Kang, J.: Ranking paragraphs for improving answer recall in open-domain question answering. In: Riloff, E., Chiang, D., Hockenmaier, J., Tsujii, J. (eds.) Proceedings of the 2018 EMNLP. pp. 565--569. Association for Computational Linguistics, Brussels, Belgium (Oct-Nov 2018). \doi{10.18653/v1/D18-1053}

\bibitem{lee-etal-2019-latent}
Lee, K., Chang, M.W., Toutanova, K.: Latent retrieval for weakly supervised open domain question answering. In: Proceedings of the 57th Annual Meeting of the Association for Computational Linguistics. pp. 6086--6096. Association for Computational Linguistics, Florence, Italy (Jul 2019). \doi{10.18653/v1/P19-1612}, \url{https://aclanthology.org/P19-1612}

\bibitem{liEtAl-2021-semantic}
Li, Z., Tomar, Y., Passonneau, R.J.: A semantic feature-wise transformation relation network for automatic short answer grading. In: Moens, M.F., Huang, X., Specia, L., Yih, S.W.t. (eds.) Proceedings of the 2021 {EMNLP}. pp. 6030--6040. Association for Computational Linguistics, Online and Punta Cana, Dominican Republic (Nov 2021). \doi{10.18653/v1/2021.emnlp-main.487}

\bibitem{mathias&bhattacharyya_asap_lrec18}
Mathias, S., Bhattacharyya, P.: {ASAP}++: Enriching the {ASAP} automated essay grading dataset with essay attribute scores. In: Proceedings of the Eleventh International Conference on Language Resources and Evaluation ({LREC} 2018). European Language Resources Association (ELRA), Miyazaki, Japan (2018), \url{https://aclanthology.org/L18-1187}

\bibitem{mizumoto-etal-2019-analytic}
Mizumoto, T., Ouchi, H., Isobe, Y., Reisert, P., Nagata, R., Sekine, S., Inui, K.: Analytic score prediction and justification identification in automated short answer scoring. In: Proceedings of the Fourteenth Workshop on Innovative Use of NLP for Building Educational Applications. pp. 316--325. Association for Computational Linguistics, Florence, Italy (Aug 2019). \doi{10.18653/v1/W19-4433}, \url{https://aclanthology.org/W19-4433}

\bibitem{o2016scholarly}
O’Donovan, B., Rust, C., Price, M.: A scholarly approach to solving the feedback dilemma in practice. Assessment \& Evaluation in Higher Education  \textbf{41}(6),  938--949 (2016)

\bibitem{panadero-et-al2023}
Panadero, E., Jonsson, A., Pinedo, L., Fern\'{a}ndez-Castilla, B.: Effects of rubrics on academic performance, self-regulated learning, and self-efficacy: a meta-analytic review. Educational Psychology Review  \textbf{35, article 113} (2023). \doi{10.1007/s10648-023-09823-4}

\bibitem{data}
Passonneau, R.J., Li, Z., Atil, B., Koenig, K.M.: Reliable rubric-based assessment of physics lab reports: Data for machine learning (2022). \doi{10.26208/BWE2-BR31}

\bibitem{passonneau-masi-lrec06}
Passonneau, R.J.: Measuring agreement on set-valued items ({MASI}) for semantic and pragmatic annotation. In: Proceedings of the 5th International Conference on Language Resources and Evaluation ({LREC}{'}06). ELRA, Genoa, Italy (2006)

\bibitem{becky_data}
Passonneau, R.J., Koenig, K., Li, Z., Soddano, J.: The ideal versus the real deal in assessment of physics lab report writing. European Journal of Applied Sciences  \textbf{11}(2),  626–644 (Apr 2023). \doi{10.14738/aivp.112.14406}

\bibitem{puntambekarEtAl_science-explanations_icls23}
Puntambekar, S., Dey, I., Gnesdilow, D., Passonneau, R.J., Kim, C.: Examining the effect of automated assessments and feedback on students’ written science explanations. In: Blikstein, P., Van~Aalst, J., Kizito, R., Brennan, K. (eds.) 17th International Conference of the Learning Sciences (ICLS 2023). pp. 1865--1866. International Society of the Learning Sciences (2023), \url{https://repository.isls.org//handle/1/10060}

\bibitem{rahimiEtAl17}
Rahimi, Z., Litman, D.J., Correnti, R., Wang, E., Matsumura, L.C.: Assessing students' use of evidence and organization in response-to-text writing: Using natural language processing for rubric-based automated scoring. Int. J. Artif. Intell. Educ.  \textbf{27}(4),  694--728 (2017)

\bibitem{reimers-gurevych-2019-sentence}
Reimers, N., Gurevych, I.: Sentence-{BERT}: Sentence embeddings using {S}iamese {BERT}-networks. In: Proceedings of the 2019 Conference on Empirical Methods in Natural Language Processing and the 9th International Joint Conference on Natural Language Processing (EMNLP-IJCNLP). pp. 3982--3992. Association for Computational Linguistics, Hong Kong, China (Nov 2019). \doi{10.18653/v1/D19-1410}, \url{https://aclanthology.org/D19-1410}

\bibitem{Ridley_He_Dai_Huang_Chen_2021}
Ridley, R., He, L., Dai, X.y., Huang, S., Chen, J.: Automated cross-prompt scoring of essay traits. Proceedings of the AAAI Conference on Artificial Intelligence  \textbf{35}(15),  13745--13753 (May 2021). \doi{10.1609/aaai.v35i15.17620}, \url{https://ojs.aaai.org/index.php/AAAI/article/view/17620}

\bibitem{sachanEtAl-2021-end}
Sachan, D., Patwary, M., Shoeybi, M., Kant, N., Ping, W., Hamilton, W.L., Catanzaro, B.: End-to-end training of neural retrievers for open-domain question answering. In: Zong, C., Xia, F., Li, W., Navigli, R. (eds.) Proceedings of the 59th {ACL} and the 11th {IJCNL}. pp. 6648--6662. ACL, Online (Aug 2021). \doi{10.18653/v1/2021.acl-long.519}

\bibitem{schick2021self}
Schick, T., Udupa, S., Sch{\"u}tze, H.: Self-diagnosis and self-debiasing: A proposal for reducing corpus-based bias in {NLP}. Transactions of the ACL  \textbf{9},  1408--1424 (2021)

\bibitem{shibata&uto-2022-analytic}
Shibata, T., Uto, M.: Analytic automated essay scoring based on deep neural networks integrating multidimensional item response theory. In: Proceedings of the 29th {ICCL}. pp. 2917--2926. International Committee on Computational Linguistics, Gyeongju, Republic of Korea (Oct 2022), \url{https://aclanthology.org/2022.coling-1.257}

\bibitem{singh2021end}
Singh, D., Reddy, S., Hamilton, W., Dyer, C., Yogatama, D.: End-to-end training of multi-document reader and retriever for open-domain question answering. Advances in Neural Information Processing Systems  \textbf{34},  25968--25981 (2021)

\bibitem{singh_automated_2022}
Singh, P., Passonneau, R.J., Wasih, M., Cang, X., Kim, C., Puntambekar, S.: Automated {Support} to {Scaffold} {Students}’ {Written} {Explanations} in {Science}. In: Rodrigo, M.M., Matsuda, N., Cristea, A.I., Dimitrova, V. (eds.) Artificial {Intelligence} in {Education}, vol. 13355, pp. 660--665. Springer (2022). \doi{10.1007/978-3-031-11644-5\_64}

\bibitem{sungEtAl-2019-pre}
Sung, C., Dhamecha, T., Saha, S., Ma, T., Reddy, V., Arora, R.: Pre-training {BERT} on domain resources for short answer grading. In: Proceedings of the 2019 {EMNLP} and the 9th {IJCNLP}. pp. 6071--6075. Association for Computational Linguistics, Hong Kong, China (Nov 2019). \doi{10.18653/v1/D19-1628}

\bibitem{takano&ichikawa-2022-automatic}
Takano, S., Ichikawa, O.: Automatic scoring of short answers using justification cues estimated by {BERT}. In: Kochmar, E., Burstein, J., Horbach, A., Laarmann-Quante, R., Madnani, N., Tack, A., Yaneva, V., Yuan, Z., Zesch, T. (eds.) Proceedings of the 17th {BEA} Workshop. pp. 8--13. Association for Computational Linguistics, Seattle, Washington (Jul 2022). \doi{10.18653/v1/2022.bea-1.2}

\bibitem{wang2021data}
Wang, T., Funayama, H., Ouchi, H., Inui, K.: Data augmentation by rubrics for short answer grading. Journal of Natural Language Processing  \textbf{28}(1),  183--205 (2021)

\bibitem{wangEtAl-naacl22}
Wang, Y., Wang, C., Li, R., Lin, H.: On the use of {BERT} for automated essay scoring: Joint learning of multi-scale essay representation. In: Proceedings of the 2022 Conference of the North American Chapter of the ACL ({NAACL}). pp. 3416--3425. Association for Computational Linguistics (2022). \doi{10.18653/v1/2022.naacl-main.249}

\bibitem{xie-etal-2022-automated}
Xie, J., Cai, K., Kong, L., Zhou, J., Qu, W.: Automated essay scoring via pairwise contrastive regression. In: Proceedings of the 29th International Conference on Computational Linguistics. pp. 2724--2733. International Committee on Computational Linguistics, Gyeongju, Republic of Korea (Oct 2022), \url{https://aclanthology.org/2022.coling-1.240}

\bibitem{yangEtAl-2020-enhancing}
Yang, R., Cao, J., Wen, Z., Wu, Y., He, X.: Enhancing automated essay scoring performance via fine-tuning pre-trained language models with combination of regression and ranking. In: Findings of EMNLP 2020. pp. 1560--1569. ACL, Online (Nov 2020). \doi{10.18653/v1/2020.findings-emnlp.141}

\end{thebibliography}

\newpage

\end{document}


\section*{Rubrics and Score Distributions}

\begin{figure*}[ht]
    \centering
    \includegraphics[scale=0.3]{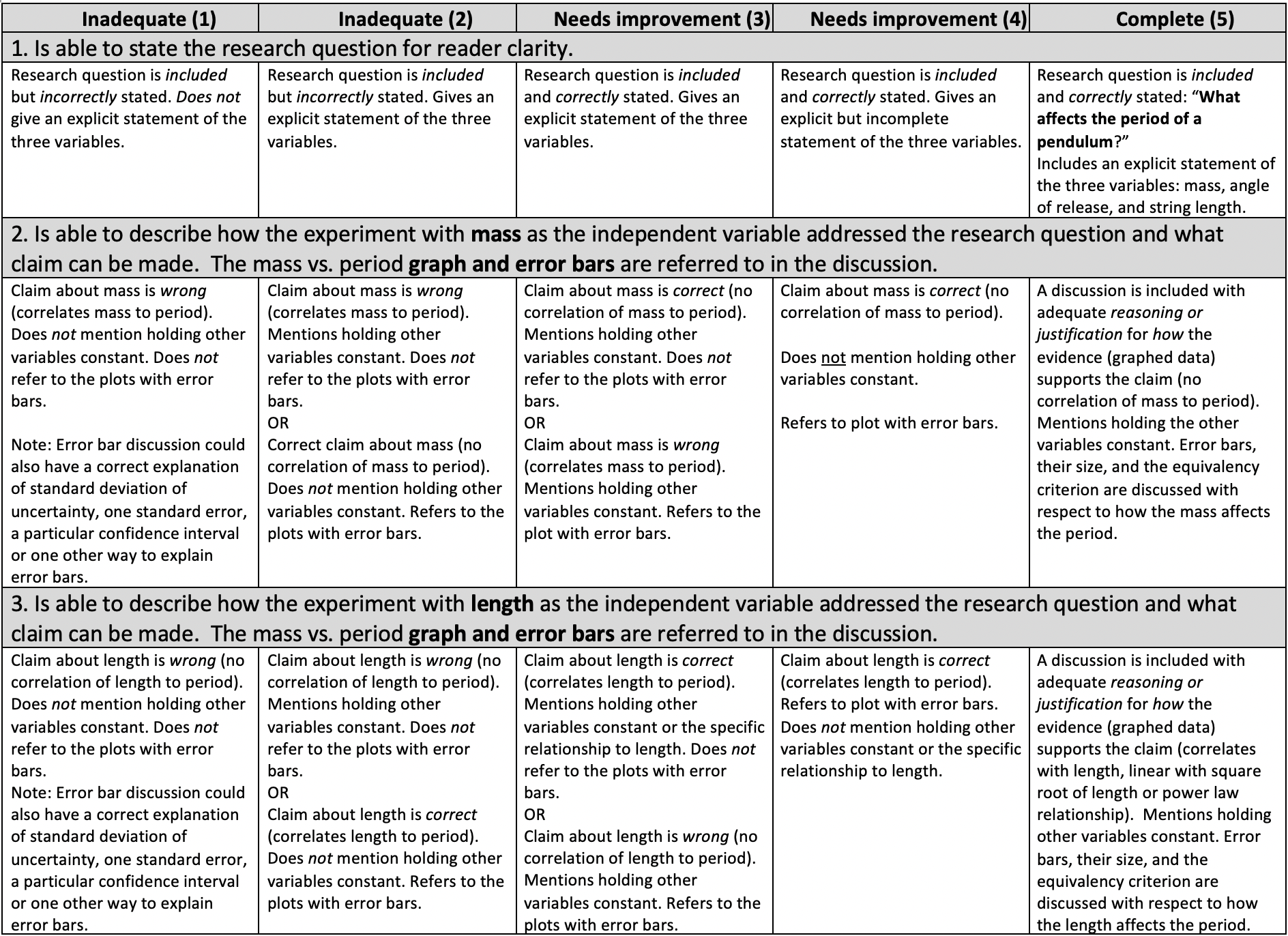}
    \includegraphics[scale=0.3]{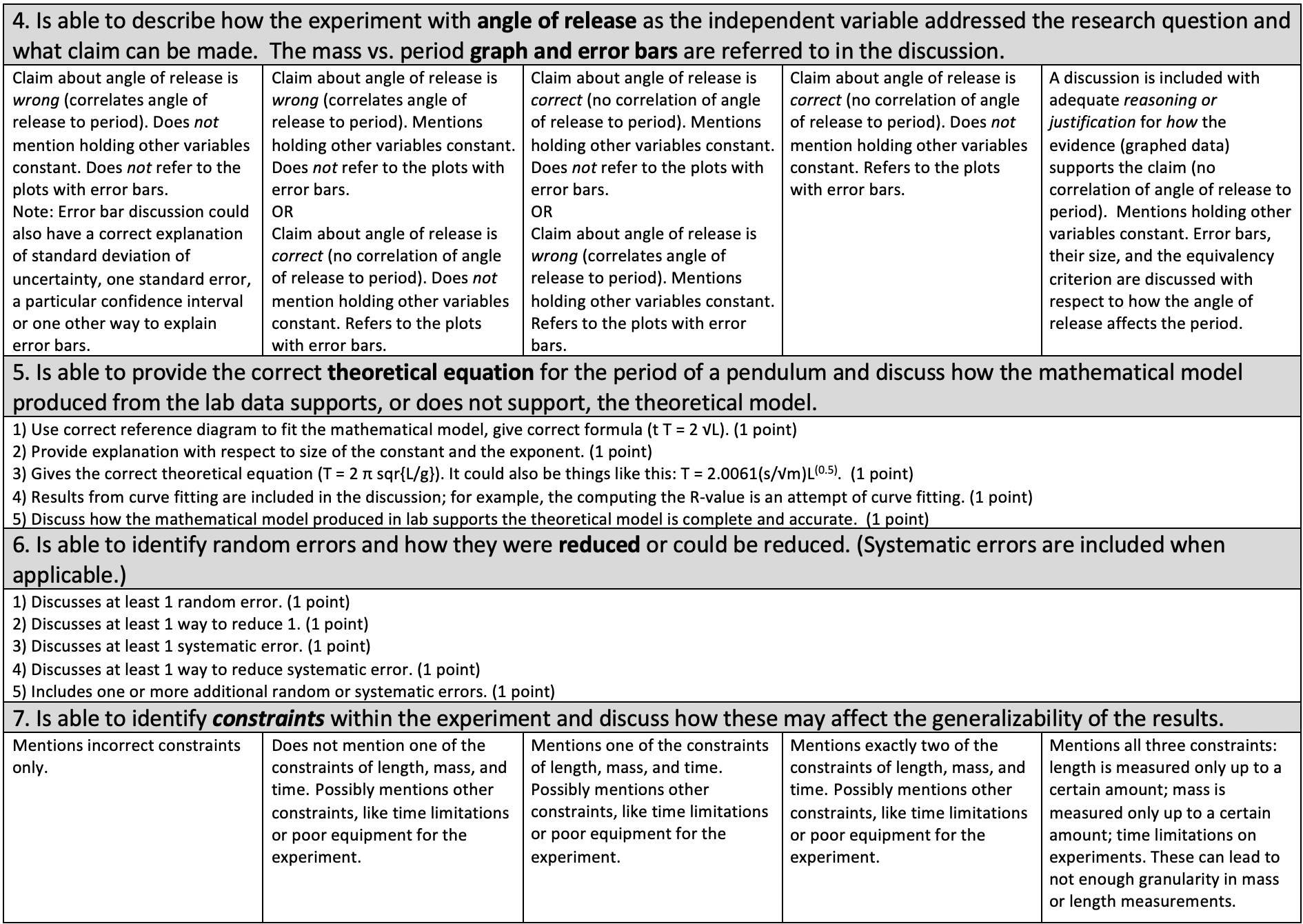}
   \caption{Pendulum Report Analytic Rubric}
    \label{fig:rubric-pendulum}
\end{figure*}

\label{app:rubric}
\begin{figure*}[ht]
\centering
    \includegraphics[scale=0.3]{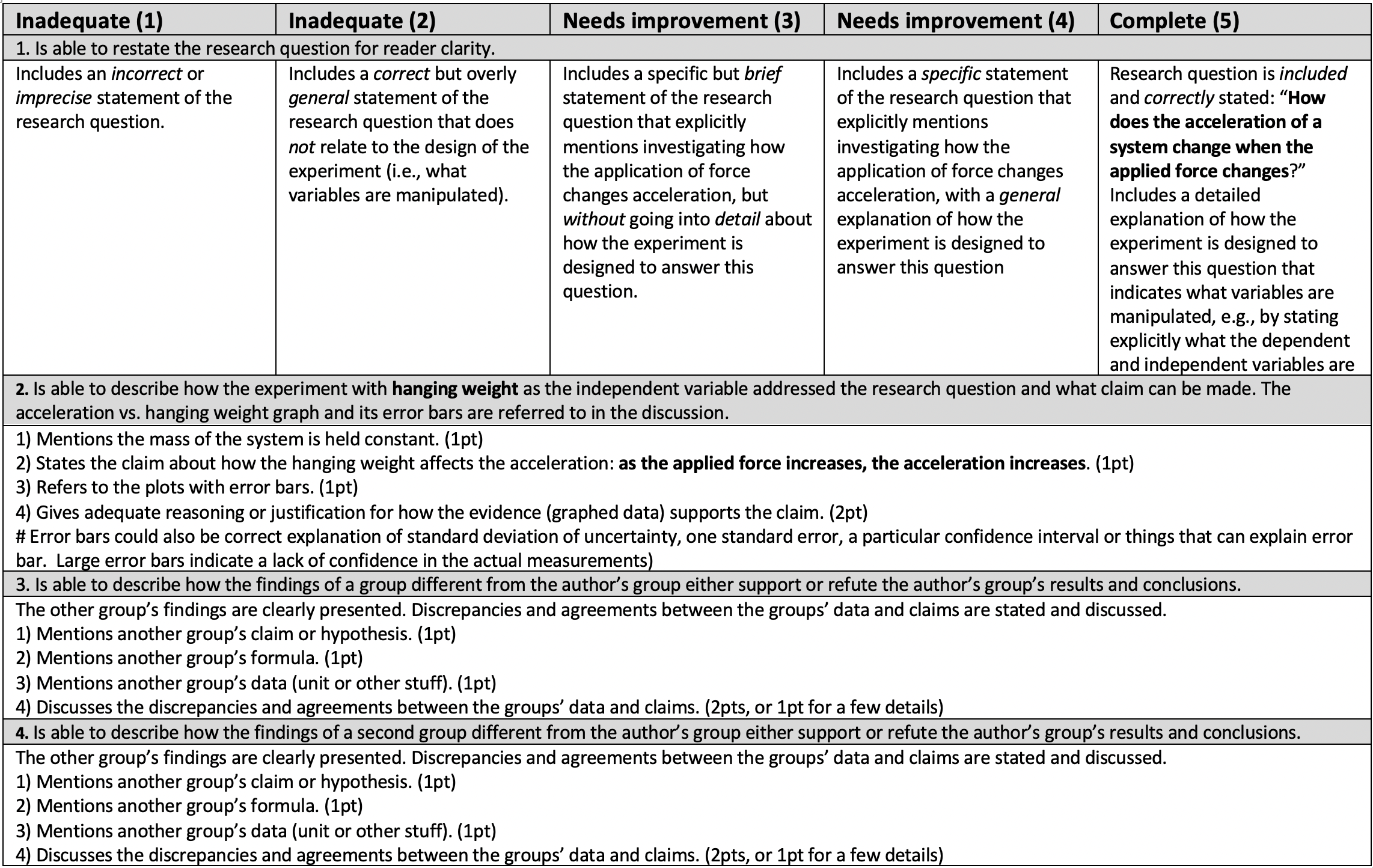}
    \includegraphics[scale=0.354]{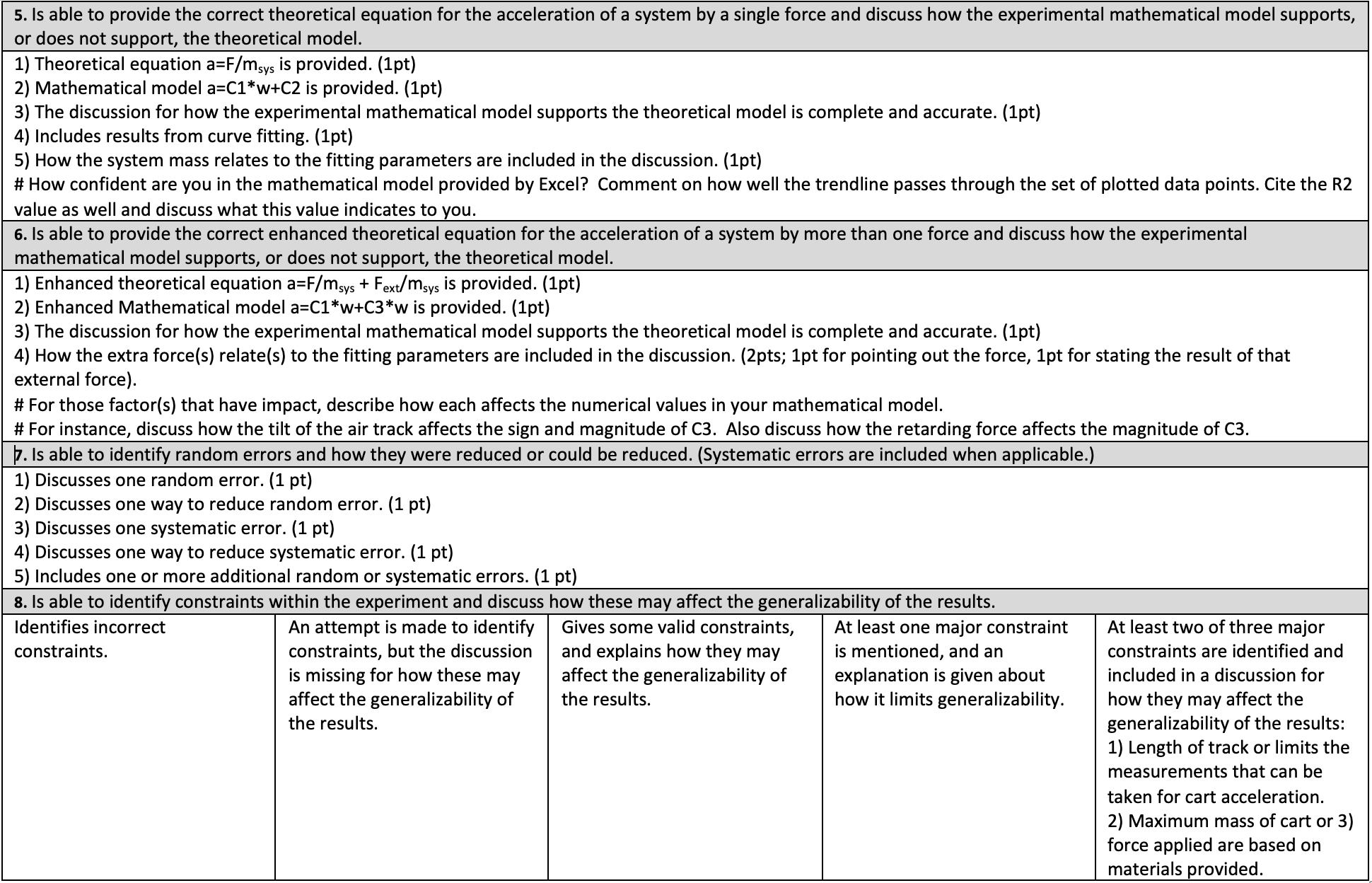}\caption{Newton’s Second Law Analytic Rubric}
\label{fig:rubric-newton}
\end{figure*}


\begin{figure*}
\centering
\begin{subfigure}{.5\textwidth}
  \centering
  \includegraphics[width=.9\linewidth]{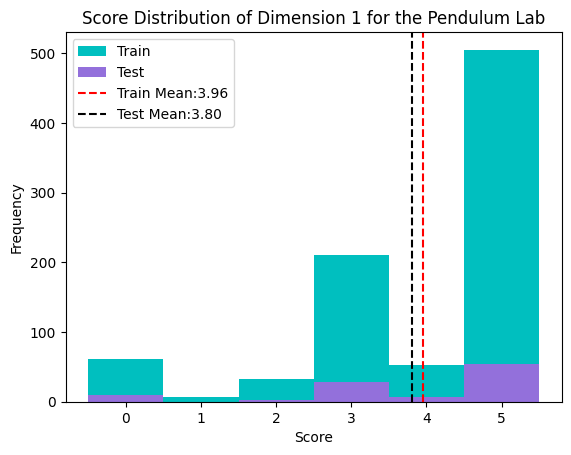}
  \caption{}
\end{subfigure}%
\begin{subfigure}{.5\textwidth}
  \centering
  \includegraphics[width=.9\linewidth]{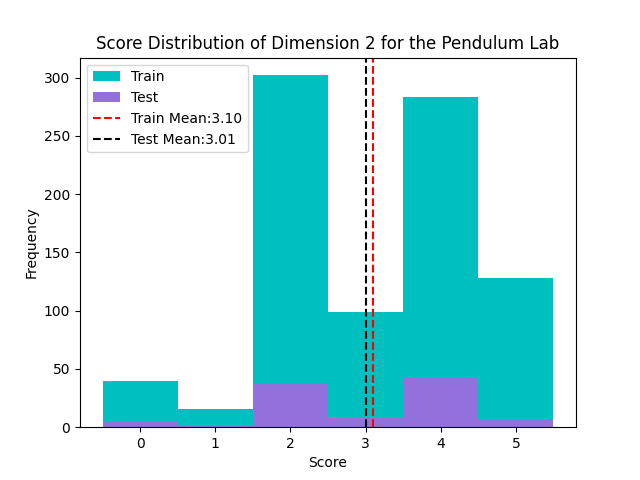}
  \caption{}
\end{subfigure}

\begin{subfigure}{.5\textwidth}
  \centering
  \includegraphics[width=.9\linewidth]{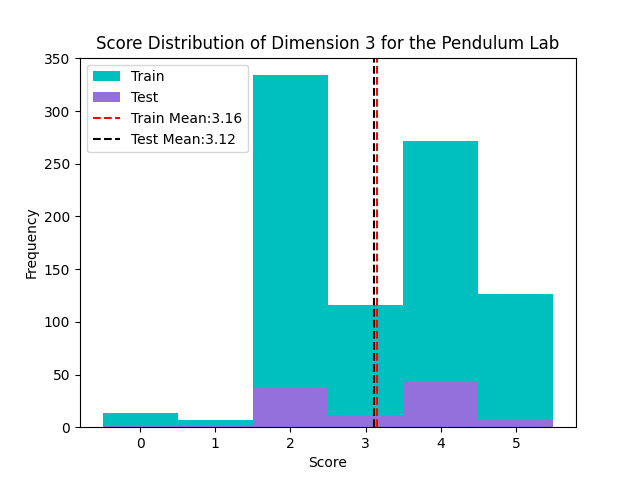}
  \caption{}
\end{subfigure}%
\begin{subfigure}{.5\textwidth}
  \centering
  \includegraphics[width=.9\linewidth]{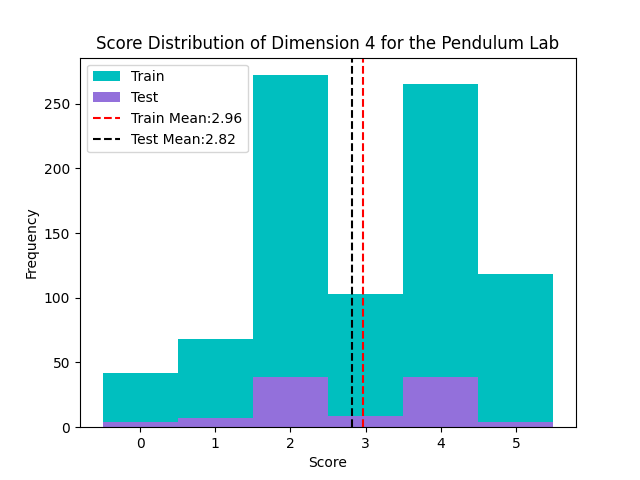}
  \caption{}
\end{subfigure}

\begin{subfigure}{.5\textwidth}
  \centering
  \includegraphics[width=.9\linewidth]{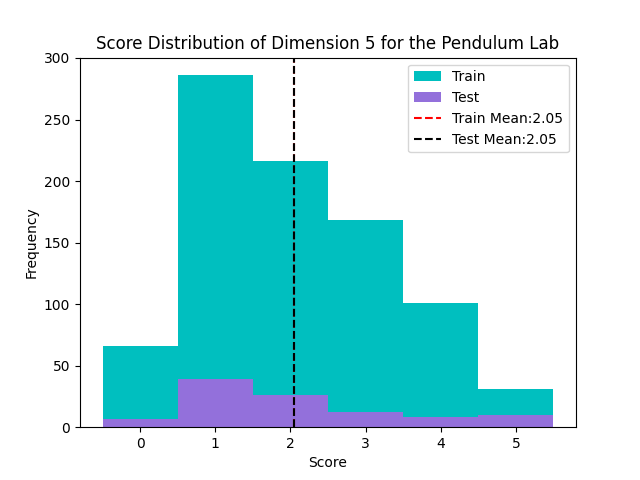}
  \caption{}
\end{subfigure}%
\begin{subfigure}{.5\textwidth}
  \centering
  \includegraphics[width=.9\linewidth]{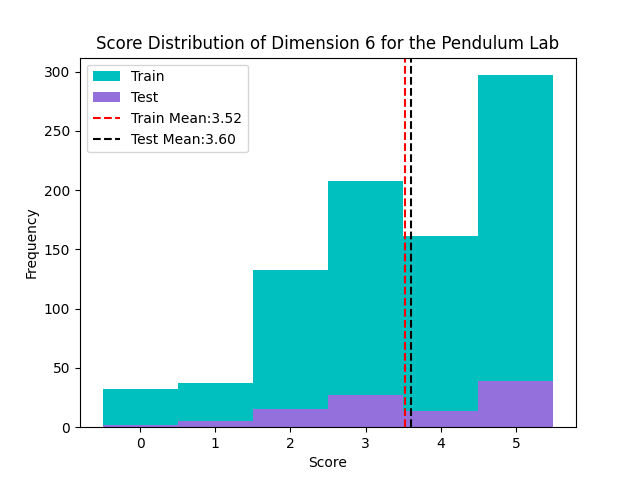}
  \caption{}
\end{subfigure}

\begin{subfigure}{.5\textwidth}
  \centering
  \includegraphics[width=.7\linewidth]{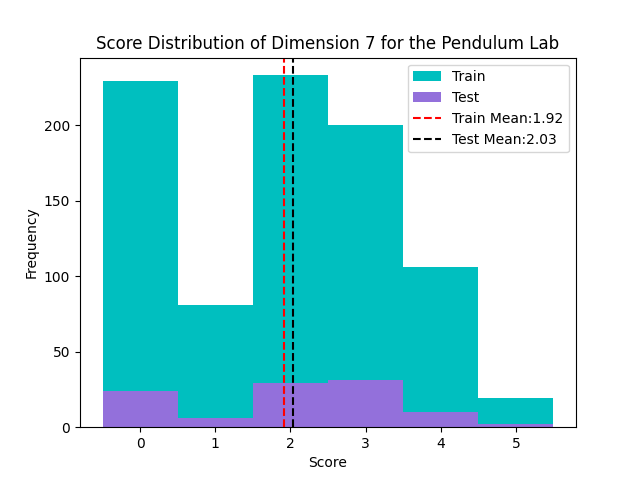}
  \caption{}
\end{subfigure}%
\caption{Score distributions per dimension for the first lab.}
\label{fig:All_dimension_score_dist_pendulum}
\end{figure*}

\begin{figure*}
\centering
\begin{subfigure}{.5\textwidth}
  \centering
  \includegraphics[width=.9\linewidth]{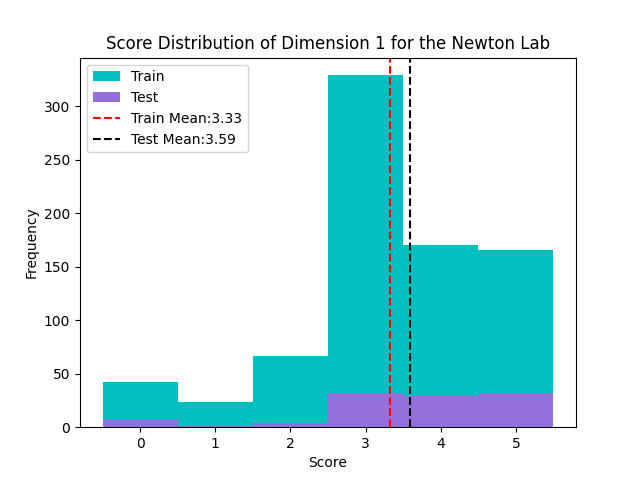}
  \caption{}
\end{subfigure}%
\begin{subfigure}{.5\textwidth}
  \centering
  \includegraphics[width=.9\linewidth]{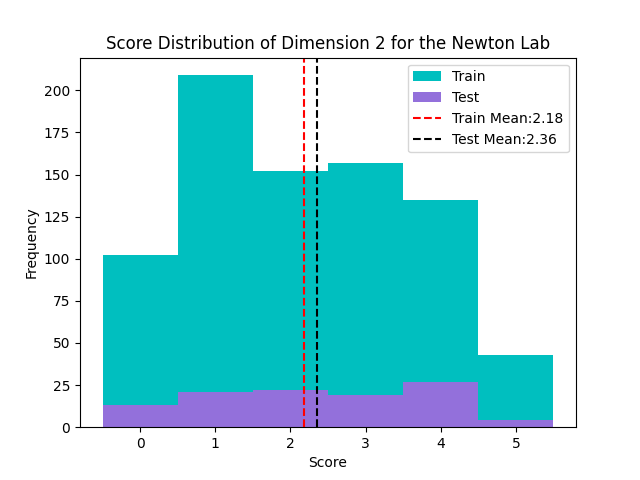}
  \caption{}
\end{subfigure}

\begin{subfigure}{.5\textwidth}
  \centering
  \includegraphics[width=.9\linewidth]{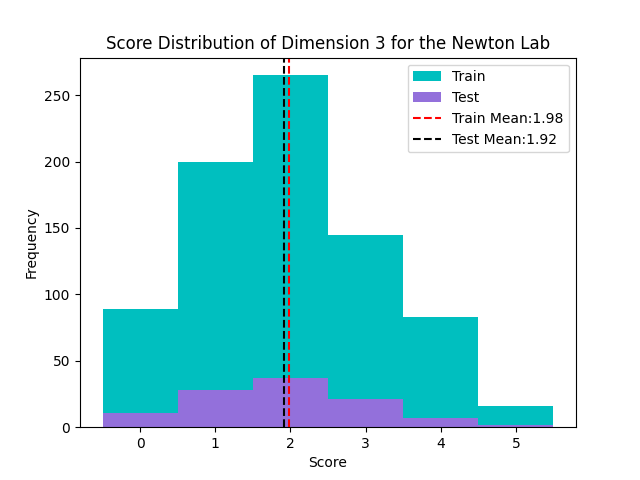}
  \caption{}
\end{subfigure}%
\begin{subfigure}{.5\textwidth}
  \centering
  \includegraphics[width=.9\linewidth]{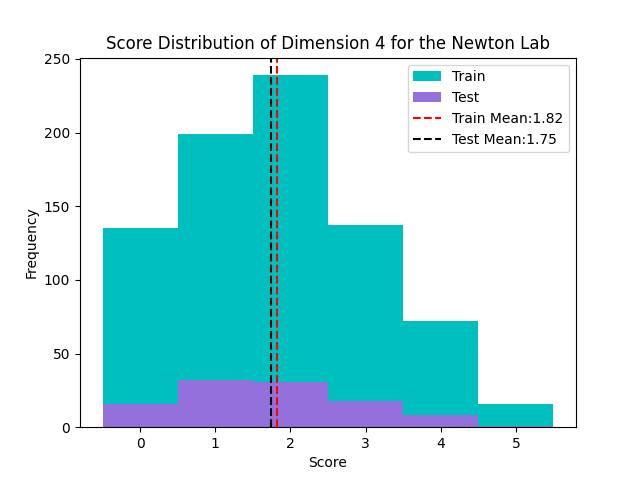}
  \caption{}
\end{subfigure}

\begin{subfigure}{.5\textwidth}
  \centering
  \includegraphics[width=.9\linewidth]{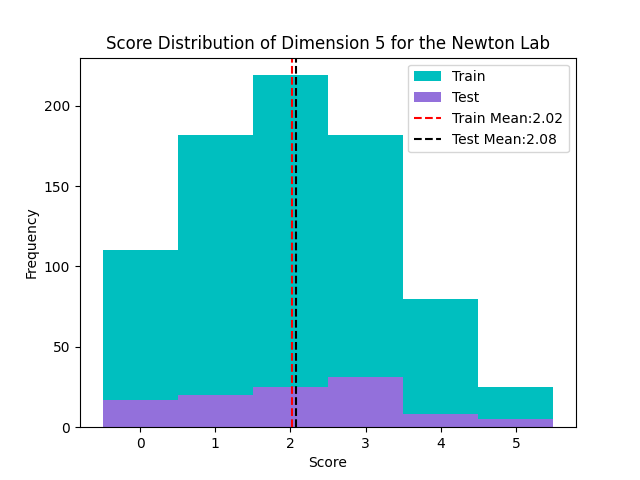}
  \caption{}
\end{subfigure}%
\begin{subfigure}{.5\textwidth}
  \centering
  \includegraphics[width=.9\linewidth]{figures/descriptive_stats/newton_dim6_score.png}
  \caption{}
\end{subfigure}

\begin{subfigure}{.5\textwidth}
  \centering
  \includegraphics[width=.8\linewidth]{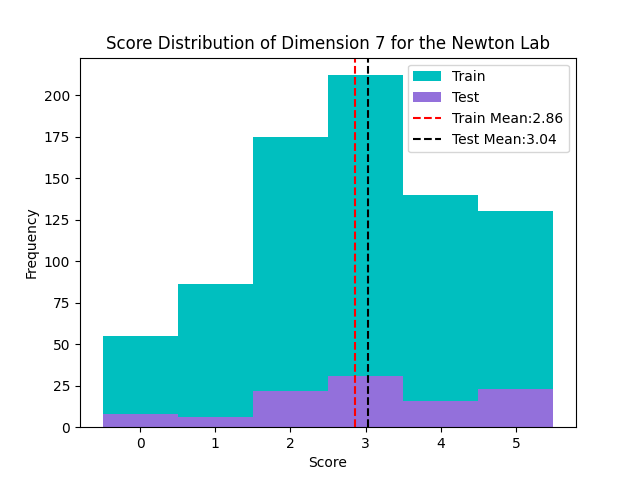}
  \caption{}
\end{subfigure}%
\begin{subfigure}{.5\textwidth}
  \centering
  \includegraphics[width=.8\linewidth]{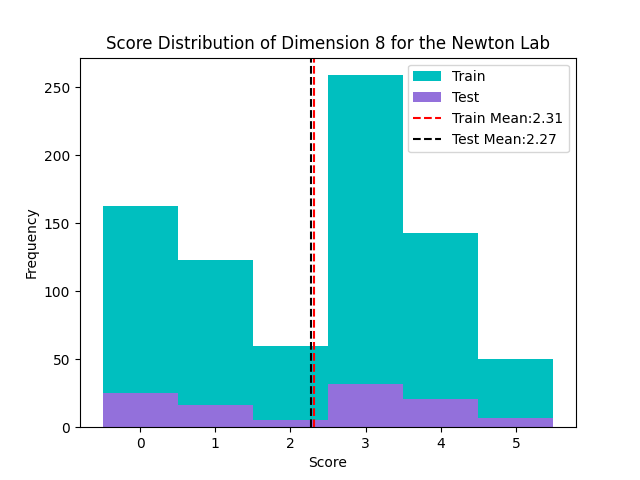}
  \caption{}
\end{subfigure}
\caption{Score distributions per dimension for the second lab.}
\label{fig:All_dimension_score_dist_newton}
\end{figure*}